\documentclass[11pt]{article}
\usepackage{enumitem}

\usepackage[final]{acl}

\usepackage{times}
\usepackage{booktabs}
\usepackage{multirow}
\usepackage{latexsym}
\usepackage{pifont}
\usepackage{amsfonts}
\usepackage{amssymb}
\usepackage{tcolorbox}
\usepackage{amsmath}
\usepackage{hyperref}
\usepackage[T1]{fontenc}

\usepackage[utf8]{inputenc}

\usepackage{microtype}

\usepackage{inconsolata}

\usepackage{graphicx}
\usepackage{dblfloatfix}

\title{SPARROW: Scalable Taxonomy Induction via Structure-Preserving Partitioning and Constraint-Guided Merging}

\author{
Yirui Zhang \quad Yixuan Tang\thanks{Corresponding author.} \quad Yandong Sun \quad Mong-Li Lee \quad Anthony Kum Hoe Tung \\
School of Computing, National University of Singapore \\
\texttt{\{yirui\_z, yixuan, yandong, leeml, atung\}@comp.nus.edu.sg}
}

\begin{document}
\maketitle


\begin{abstract}
Taxonomy induction aims to organize concept sets into coherent hierarchical structures. Recent LLM-based methods can induce taxonomies directly from flat term lists, avoiding the need for corpora, but degrade sharply as concept sets scale up. We argue that this degradation stems not only from context length limitations, but also from structural failures in hierarchical reasoning. To address this, we adopt a divide-and-merge paradigm that partitions concepts into smaller subsets, induces local taxonomies, and merges them into a global hierarchy. However, we identify two structural failure modes inherent to this paradigm: \textbf{Structural Fragmentation}, where partitioning weakens local hierarchical signals, and \textbf{Parent Displacement}, where locally plausible relations are misplaced in the global hierarchy. To address both, we propose \textbf{SPARROW}, a scalable taxonomy induction framework that combines structure-preserving spectral partitioning to retain hierarchical connectivity within each block, and constraint-guided incremental fusion that treats block-level relations as structural constraints rather than ground truth for global placement. Experiments on large-scale benchmarks show that SPARROW consistently achieves the strongest global structural quality across backbones. The code is available at \url{https://github.com/rebeccazyr/SPARROW}.
\end{abstract}



\section{Introduction}

\begin{figure}[t]
    \centering
    \includegraphics[width=\columnwidth]{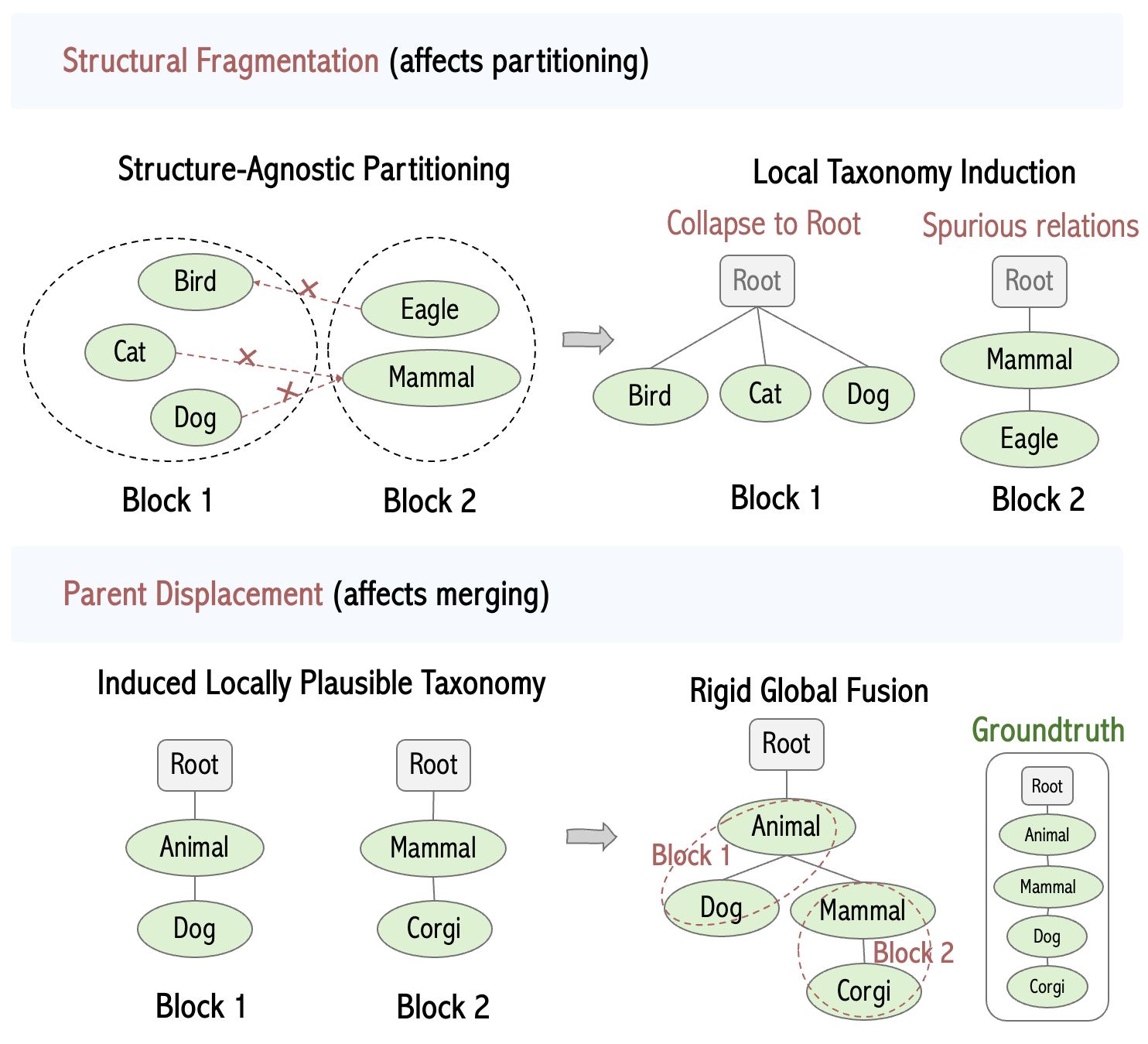}
    \caption{Two structural failure modes.}
    \label{fig:types}
\end{figure}

Taxonomies are fundamental structures for organizing concepts and supporting downstream applications such as semantic search~\cite{DBLP:conf/emnlp/KangZJ00Y24}, recommendation~\cite{DBLP:conf/icde/TanYWCLZ22}, question answering~\cite{DBLP:journals/corr/abs-2603-09341,sun2026distilling}, and, more broadly, knowledge integration~\cite{tung2026deepconnect}. Taxonomy induction has therefore been widely studied, with approaches ranging from pattern-based \cite{DBLP:conf/acl/RollerKN18} and topic model-based methods \cite{DBLP:journals/jacm/BleiGJ10} to embedding-based techniques \cite{DBLP:conf/kdd/ZhangTCSJSV018,DBLP:conf/www/ShangZLL020}, all of which rely on large text corpora to surface hierarchical signals. Recent work has shown that LLMs, by encoding rich world knowledge during pretraining, can infer hierarchical relations directly from term names \cite{DBLP:conf/models/ChenYV23,DBLP:conf/cikm/0001BTFLZ024}, achieving strong performance on small to medium-sized concept sets without corpus dependency. However, their performance deteriorates sharply as the concept set scales up. \citet{DBLP:conf/cikm/0001BTFLZ024} show that even state-of-the-art LLM-based methods lose the ability to faithfully cover all input concepts and exhibit consistent structural degradation as the term set grows, revealing fundamental limitations in both context handling capacity and hierarchical reasoning ability. More fundamentally, we argue that this degradation reflects an inherent difficulty of LLM-based approaches: maintaining globally coherent hierarchical relations under localized reasoning becomes increasingly difficult as scale grows, even when individual placements appear locally correct.

A natural response to this limitation is to adopt a divide-and-merge strategy: partition the concept set into smaller subsets, induce local taxonomies independently, and merge the resulting structures into a global hierarchy. However, a naive application of this paradigm can be brittle at scale because it overlooks two structural failure modes, illustrated in Figure~\ref{fig:types}.

\textbf{Structural Fragmentation.} Partitioning inevitably severs cross-subset parent-child relations, but the critical failure mode lies in partition quality. When concepts within a block share insufficient latent hierarchical structure, local induction tends to collapse concepts onto the root, offloading all structural recovery to the merge stage, or fabricates spurious relations that persist irrecoverably through the pipeline. Effective partitioning should therefore group concepts likely to share genuine hierarchical relations, maximizing local inductive signal despite the true taxonomy being unknown at partition time.

\textbf{Parent Displacement.}
Even when block taxonomies are locally correct, block-level parent-child 
relations do not necessarily hold globally. Because a block contains only 
a subset of concepts, intermediate nodes absent from the block cause 
ancestor-descendant relations to compress into spurious direct edges. 
Naively lifting block taxonomies as rigid units therefore embeds these 
structural errors irrecoverably into the global hierarchy. Block-level 
relations should instead constrain rather than determine global placement, 
with ambiguous 
placements resolved through explicit reasoning over its 
candidate-specific hierarchical context.


To address these failure modes, we propose \textbf{SPARROW} (Structure-Preserving 
pARtitioning and constRaint-guided meRging for taxonOmies), a scalable taxonomy 
induction framework that follows a divide-and-merge paradigm. In the divide stage, 
structure-preserving partitioning applies spectral clustering to preserve 
hierarchical connectivity within each block, retaining latent parent-child signal 
for reliable local induction under bounded context. In the merge stage, 
constraint-guided incremental fusion treats block-level relations as structural 
constraints to scope candidate parents rather than lifting them as ground truth, 
with final placement resolved through explicit LLM reasoning.

We evaluate SPARROW on multiple taxonomy induction benchmarks with concept sets ranging from tens to ten thousand nodes, using both strong-context and limited-context backbone models. Experimental results show that SPARROW consistently achieves greater structural stability than existing methods, with particularly strong improvements as scale increases.

Our main contributions are:

\begin{itemize}[itemsep=2pt, topsep=2pt]
    \item We identify two structural failure modes in \textbf{large-scale LLM-based taxonomy induction}: \textbf{Structural Fragmentation}, where partitioning weakens local hierarchical signals, and \textbf{Parent Displacement}, where locally plausible relations lead to incorrect global ancestor placement.

    \item We propose \textbf{SPARROW}, a scalable divide-and-merge framework that combines \textbf{structure-preserving partitioning} with \textbf{constraint-guided incremental fusion}, treating block-level relations as structural constraints rather than fixed global decisions.

    \item We demonstrate that SPARROW achieves \textbf{stronger global structural consistency and scalability} across multiple taxonomy benchmarks, with consistently higher ancestor-level accuracy under both strong and limited-context LLM backbones.
\end{itemize}

\section{Related Work}

\subsection{Corpus-based Taxonomy Induction}
Taxonomy induction from text corpora exploits distributional
evidence to approximate hypernym--hyponym relations.
Early approaches rely on lexical-syntactic
patterns~\cite{DBLP:conf/acl/RollerKN18,DBLP:conf/nips/SnowJN04}
or probabilistic topic models such as
hLDA~\cite{DBLP:journals/jacm/BleiGJ10,blei2003latent}, while
clustering-based methods construct hierarchies over learned term
embeddings~\cite{DBLP:conf/kdd/ZhangTCSJSV018,
DBLP:conf/www/ShangZLL020}.
More recent work incorporates LLMs into corpus-based pipelines
to improve taxonomy
quality~\cite{DBLP:conf/acl/KarguptaZZZMH25,DBLP:conf/emnlp/LahiriHS25,DBLP:conf/emnlp/ZhuLGHFQ25,DBLP:journals/corr/abs-2601-05520}.
However, these methods assume access to document corpora and aim to
induce taxonomies from distributional evidence, where the concept inventory is
not predefined. In contrast, our setting assumes a fixed flat concept set
without accompanying text, and focuses on inferring the hierarchical
structure among the given concepts.

\begin{figure*}[t]
  \centering
  \includegraphics[width=\textwidth]{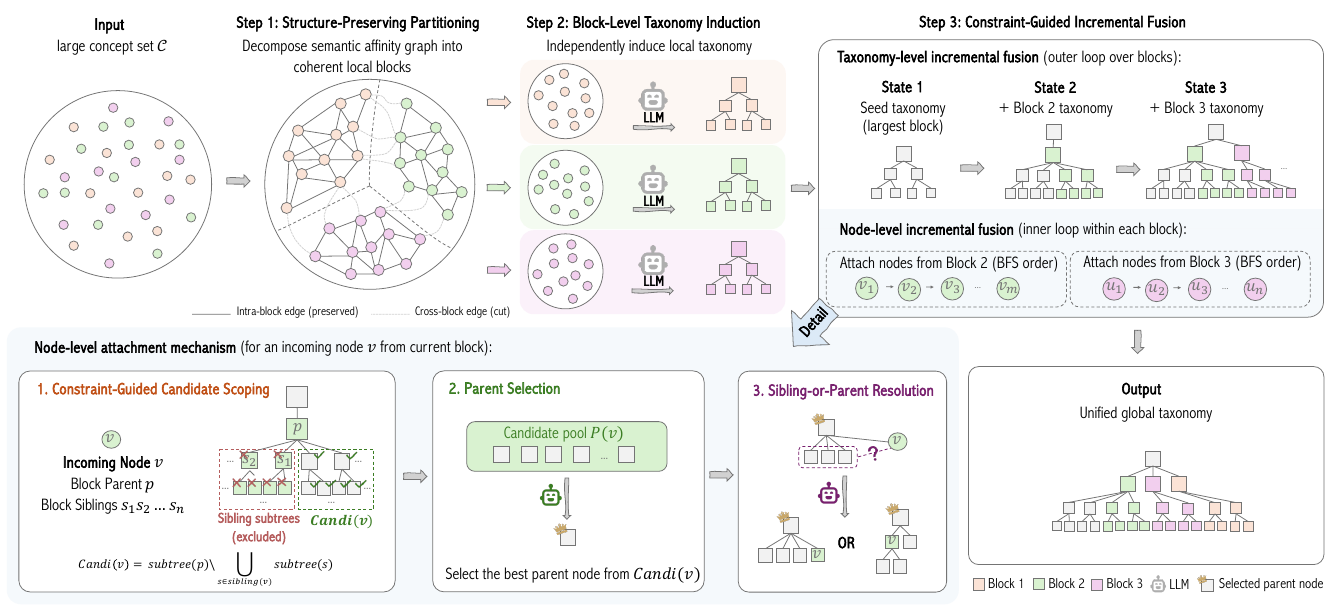}
  \caption{
Overview of \textbf{SPARROW}, a divide-and-merge framework for large-scale taxonomy induction.
SPARROW first partitions the input concepts into semantically coherent blocks through \emph{Structure-Preserving Partitioning}, and induces local taxonomies through \emph{Block-Level Taxonomy Induction}.
The resulting block-level taxonomies are then fused into a unified global taxonomy through \emph{Constraint-Guided Incremental Fusion}, which consists of an outer loop that sequentially inserts local taxonomies into the evolving global structure, and an inner loop that attaches incoming nodes under structural constraints.
}
  \label{fig:main}
\end{figure*}

\subsection{Term-based Taxonomy Induction}
The emergence of large language models has
enabled taxonomy induction directly from flat
concept sets by leveraging strong semantic
abstraction capabilities.
Prior work explores prompting and fine-tuning
strategies for hierarchical
construction~\cite{DBLP:conf/models/ChenYV23},
and layer-wise prompting schemes to improve
structural consistency~\cite{DBLP:conf/cikm/0001BTFLZ024}.
However, these approaches require all candidate
terms to be present within a single inference
context, whether generating the taxonomy in one
pass or layer by layer.
Such designs are inherently limited by context
length constraints, making it difficult to maintain
hierarchical consistency as the concept set grows.

\subsection{Taxonomy Merging and Structural Fusion}
Taxonomy merging integrates multiple hierarchical structures into a unified taxonomy~\cite{cheng2023systematic}. Existing methods based on semantic node matching~\cite{DBLP:journals/is/RaunichR14}, logic-based consistency enforcement~\cite{DBLP:journals/corr/ChenYFBL14}, and heuristic conflict resolution~\cite{DBLP:journals/jis/ChenWYLCH22,DBLP:journals/ao/BabalouK23} all presuppose that input taxonomies share overlapping nodes or explicit cross-taxonomy correspondences to anchor the integration. This condition does not hold when merging independently induced partial hierarchies with disjoint vocabularies.

Taxonomy expansion and completion methods~\cite{DBLP:conf/kdd/YuLSFSZ20,DBLP:journals/corr/abs-2406-17739,DBLP:conf/acl/MishraA025,DBLP:conf/www/WangZCZL21} insert individual concepts into a fixed backbone by selecting appropriate parent or child nodes. In contrast, our fusion procedure attaches structured block-level subtrees onto an evolving, incomplete backbone, where the internal relations within each subtree constrain attachment decisions and the backbone itself grows incrementally throughout the process.

\section{Problem Formulation}
We study the single-parent setting, where each non-root concept has exactly one parent, and represent a taxonomy as a rooted tree $\mathcal{T} = (\mathcal{V}, \mathcal{E})$, where
$\mathcal{V}$ denotes a set of concepts and $\mathcal{E} \subseteq \mathcal{V} \times \mathcal{V}$
denotes \textit{is-a} relations. Given a flat concept set $\mathcal{C} = \{c_1, \ldots, c_N\}$
as input, taxonomy induction aims to infer a valid edge set $\mathcal{E}$ such that
$(\mathcal{C}, \mathcal{E})$ forms a valid rooted tree. We focus on the large-scale setting
where $|\mathcal{C}|$ is too large for joint inference.

\section{Methodology}

\subsection{Overview}

Given a large concept set $\mathcal{C}$, SPARROW constructs a globally coherent taxonomy by
explicitly decoupling \emph{local semantic abstraction} from \emph{global structural reasoning}.
The framework follows a divide-and-merge paradigm, operationalized through three tightly coupled stages: (1) \emph{structure-preserving partitioning}, which decomposes $\mathcal{C}$ into locally coherent blocks; (2) \emph{block-level taxonomy induction}, which independently induces a local taxonomy within each block under bounded context; and (3) \emph{constraint-guided incremental fusion}, which integrates the resulting block-level taxonomies into a unified global hierarchy under structural constraints.
Figure~\ref{fig:main} provides an overview of the complete SPARROW pipeline.
This design directly targets the two structural failure modes identified earlier:
\emph{Structural Fragmentation} during partitioning and \emph{Parent Displacement}
during merging.

\subsection{Structure-Preserving Partitioning}

The \emph{Structure-Preserving Partitioning} stage decomposes the concept set $\mathcal{C}$ into subsets of optimal size within LLM context budgets, while preserving coherent local sub-hierarchies. Gold subtrees typically form connected neighborhoods in the embedding $k$NN graph, yet are only moderately concentrated around a shared centroid (Appendix~\ref{app:embedding_connectivity}). An effective partitioning strategy should therefore preserve local connectivity rather than optimize geometric compactness alone. This motivates spectral clustering, which can retain connected semantic structure within each block.

Concretely, each concept $c_i \in \mathcal{C}$ is encoded into dense semantic embeddings $\{e_i\}_{i=1}^{n}$. A sparse affinity graph is constructed over the top-$k$ nearest neighbors of each concept. For a retained edge $(i,j)$, cosine distance is $d_{ij}=1-\frac{e_i\cdot e_j}{\lVert e_i\rVert\lVert e_j\rVert}$ and its weight is $\exp(-\gamma d_{ij}^{2})$, with $\gamma=1.0$. Normalized spectral clustering~\cite{DBLP:journals/sac/Luxburg07} is applied over this graph, with the block count $B$ selected adaptively via the eigengap heuristic within a candidate range derived from concept set size. Concepts are partitioned into $B$ disjoint semantic blocks $\{\mathcal{C}^{(1)}, \dots, \mathcal{C}^{(B)}\}$, each serving as an independent bounded-context unit for local taxonomy induction.

\subsection{Block-Level Taxonomy Induction}
\label{subsec:block}
Given a block $\mathcal{C}^{(b)}$, SPARROW independently induces a local taxonomy by prompting an LLM to assign each concept a parent from within-block candidates. If no suitable parent can be identified, the concept is attached to the block root, deferring its global placement to the fusion stage. The output is a set of locally coherent block-level taxonomies that capture fine-grained semantic structure.

\begin{figure*}[t]
  \centering
  \includegraphics[width=\textwidth]{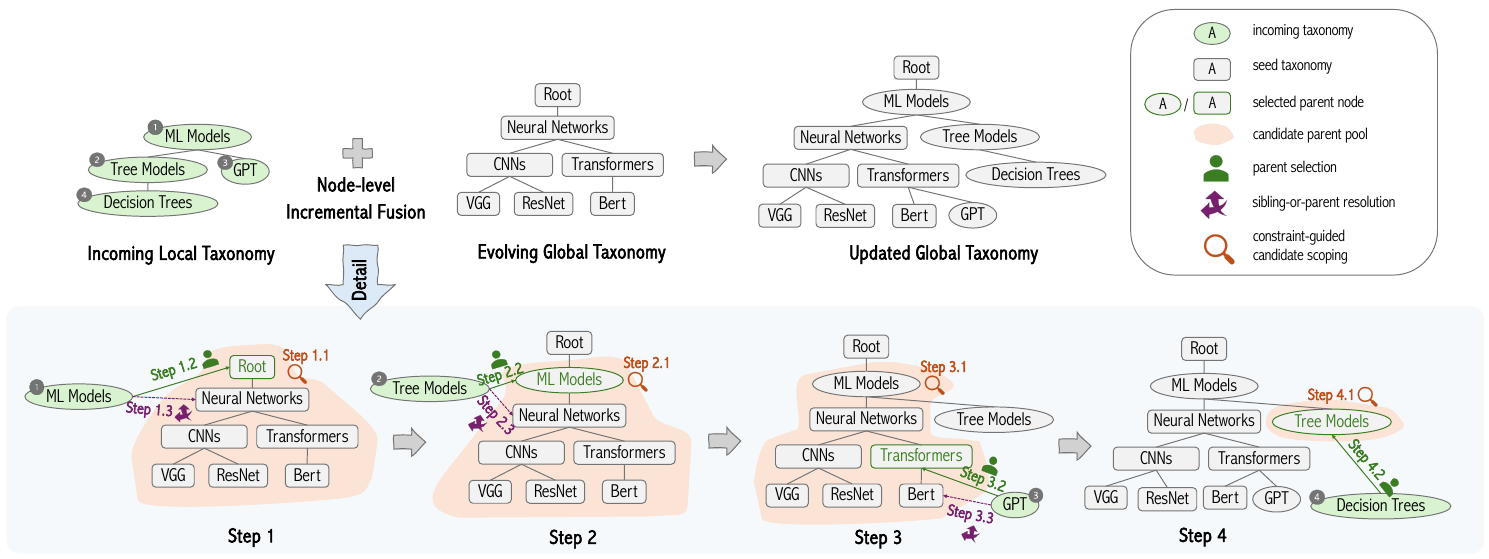}
  \caption{Running example of the \textit{Node-level Incremental Fusion} mechanism in Figure~\ref{fig:main}.
An incoming local taxonomy is integrated into the evolving global taxonomy through sequential node-level attachment. Each incoming node undergoes three stages: \textit{Constraint-Guided Candidate Scoping}, \textit{Parent Selection}, and \textit{Sibling-or-Parent Resolution}, progressively updating the global taxonomy in BFS order.}
  \label{fig:fusion}
\end{figure*}

\subsection{Constraint-Guided Incremental Fusion}
\label{subsec:psr}
\paragraph{Incremental Fusion.}
After block-level induction, SPARROW integrates the independently induced local taxonomies into a unified global hierarchy incrementally. The largest block is selected as the structural seed, maximizing initial hierarchical coverage. Remaining blocks are integrated one at a time into the evolving partial hierarchy in descending order of block size, with nodes within each block processed in BFS order so that a node's block-level parent is always placed before its descendants, enabling constraint-guided candidate scoping during attachment. Each attachment involves three sequential steps: constructing a constraint-guided candidate parent pool, selecting a valid parent, and resolving the node's structural role relative to existing children. Fusion preserves within-block ancestral relations while allowing cross-block attachments to refine direct edges.
\paragraph{Constraint-Guided Candidate Scoping.}
For each incoming node $v$ to be attached to the evolving taxonomy, a constraint-guided candidate parent pool is constructed from the currently fused taxonomy. The construction directly exploits the structural information established during block-level induction: since $v$'s parent-child and sibling relations within its block have already been determined, candidates are restricted to the subtree rooted at $v$'s block-level parent $p$, excluding the subtrees of $v$'s siblings whose parent-child relations with $v$ have already been ruled out,

\begin{equation*}
\mathrm{Candi}(v)
=
\operatorname{subtree}(p)
\setminus
\bigcup_{s\in sibling(v)}
\operatorname{subtree}(s)
\end{equation*}

For block-root nodes with no block-level parent, $\mathrm{Candi}(v) = \mathcal{T}$, the set of all currently placed nodes. This scoping preserves consistency with block-level induction, reuses its semantic reasoning, and reduces the candidate pool to a structurally relevant subset, improving both precision and efficiency of the attachment decision.

\paragraph{Parent Selection.} Given the constraint-guided candidate pool $\mathrm{Candi}(v)$, the top-$K$ candidates are first retrieved by embedding similarity and passed to an LLM for final parent selection. Each candidate is provided with its full taxonomy path and existing children, supplying the structural context necessary for coherent placement (see Appendix~\ref{parent_selection} for the full prompt). The global root is always included as a candidate, allowing the LLM to defer placement when the subsumption check finds no match. Repositioning is triggered incrementally during sibling-or-parent resolution: when a later node subsumes a deferred root child, that child is moved beneath the newly inserted node rather than being handled by a separate post-processing pass.

\paragraph{Sibling-or-Parent Resolution.} Once a parent $P$ is selected, a second LLM decision determines the structural role of $v$ relative to the existing children of $P$. If $v$'s semantic scope is comparable to that of the existing children, $v$ is attached as a sibling; if $v$ semantically subsumes any of those children, $v$ is inserted as an intermediate node above the subsumed children. (Prompts in Appendix~\ref{Sibling-or-Parent Resolution}).

Figure~\ref{fig:fusion} illustrates a complete fusion example, showing how constraint-guided candidate scoping, parent selection, and sibling-or-parent resolution interact across four sequential node attachments when integrating an incoming local taxonomy into the evolving global taxonomy.

\section{Experiments}
\subsection{Experimental Setup}

\begin{table*}[t]
\centering
\small
\setlength{\tabcolsep}{4.5pt}

\begin{tabular}{lllccccccccc}
\toprule
\textbf{Dataset} &
\textbf{Method} &
\textbf{Backbone} &
\multicolumn{3}{c}{\textbf{Node}} &
\multicolumn{3}{c}{\textbf{Edge}} &
\multicolumn{3}{c}{\textbf{Ancestor}} \\
\cmidrule(lr){4-6}
\cmidrule(lr){7-9}
\cmidrule(lr){10-12}
& & &
P & R & F1 &
P & R & F1 &
P & R & F1 \\
\midrule

\multirow{8}{*}{CCS}
& TaxoGPT
& \multirow{4}{*}{GPT-5}
& 0.984 & 0.756 & 0.855
& 0.325 & 0.242 & 0.278
& 0.696 & 0.277 & 0.397 \\
& Chain-of-Layer
&
& 1.000 & 0.519 & 0.683
& 0.600 & 0.301 & 0.401
& 0.751 & 0.165 & 0.271 \\
& LLMscorer
&
& 1.000 & 1.000 & \textbf{1.000}
& 0.490 & 0.474 & \textbf{0.482}
& 0.488 & 0.464 & 0.476 \\
& SPARROW
&
& 1.000 & 0.939 & 0.969
& 0.425 & 0.399 & 0.412
& 0.670 & 0.582 & \textbf{0.623} \\
\cmidrule(lr){2-12}
& TaxoGPT
& \multirow{4}{*}{LLaMA3-8B}
& 0.262 & 0.016 & 0.030
& 0.050 & 0.003 & 0.006
& 0.029 & 0.001 & 0.002 \\
& Chain-of-Layer
&
& -- & -- & --
& -- & -- & --
& -- & -- & -- \\
& LLMscorer
&
& 1.000 & 0.999 & \textbf{1.000}
& 0.248 & 0.207 & \textbf{0.226}
& 0.222 & 0.109 & 0.146 \\
& SPARROW
&
& 0.972 & 0.741 & 0.841
& 0.113 & 0.087 & 0.100
& 0.560 & 0.315 & \textbf{0.403} \\

\midrule

\multirow{8}{*}{Google}
& TaxoGPT
& \multirow{4}{*}{GPT-5}
& 1.000 & 0.812 & 0.896
& 0.819 & 0.432 & 0.566
& 0.700 & 0.556 & 0.619 \\
& Chain-of-Layer
&
& 1.000 & 0.438 & 0.609
& 0.636 & 0.274 & 0.383
& 0.893 & 0.246 & 0.386 \\
& LLMscorer
&
& 1.000 & 0.991 & \textbf{0.995}
& 0.597 & 0.580 & \textbf{0.588}
& 0.402 & 0.519 & 0.453 \\
& SPARROW
&
& 1.000 & 0.955 & 0.977
& 0.592 & 0.566 & 0.579
& 0.770 & 0.721 & \textbf{0.745} \\
\cmidrule(lr){2-12}
& TaxoGPT
& \multirow{4}{*}{LLaMA3-8B}
& 0.944 & 0.017 & 0.033
& 0.125 & 0.002 & 0.004
& 0.171 & 0.002 & 0.004 \\
& Chain-of-Layer
&
& -- & -- & --
& -- & -- & --
& -- & -- & -- \\
& LLMscorer
&
& 1.000 & 0.845 & \textbf{0.916}
& 0.310 & 0.214 & \textbf{0.253}
& 0.304 & 0.119 & 0.171 \\
& SPARROW
&
& 0.923 & 0.500 & 0.649
& 0.282 & 0.153 & 0.198
& 0.247 & 0.257 & \textbf{0.253} \\
\bottomrule
\end{tabular}

    \caption{Performance comparison under 1K--scale taxonomy induction settings. ``--'' indicates failure to produce a valid taxonomy output due to context limitations.}
\label{tab:main_large}
\end{table*}

\paragraph{Datasets.}
We conduct our main experiments on two taxonomies from distinct domains:

\vspace{5pt}
  \textbf{ACM Computing Classification System (CCS)}~\cite{acm_ccs_dataset}, 
an expert-curated computer science taxonomy organized into six hierarchical levels, pruned to a strict tree of 1,768 concepts via single-parent filtering.

\vspace{5pt}
 \textbf{Google Product Taxonomy (Google) \\}~\cite{google_product_taxonomy}, a large-scale taxonomy for e-commerce categorization comprising 5,595 product concepts across up to seven levels of depth.
\vspace{5pt}

Both datasets provide ground-truth hierarchical structures for quantitative evaluation, covering complementary domains of academic computing and e-commerce. To further evaluate cross-domain generalization and scalability, we additionally use SemEval TExEval-2 Food~\cite{DBLP:conf/semeval/BordeaLB16} and the biomedical Medical Subject Headings (MeSH)~\cite{lipscomb2000medical}, including a 10K-concept setting; full results for both are reported in Appendix~\ref{app:additional_results}. Sampling procedures and further details are provided in Appendix~\ref{app:data_preparation}.

\paragraph{Evaluation Metrics.}
We evaluate taxonomy quality using \textbf{Node F1}, \textbf{Edge F1}, 
and \textbf{Ancestor F1}.

\textbf{Node F1} evaluates concept coverage. A node is considered correct 
if it appears in the induced taxonomy, regardless of its structural position. 
This metric reflects whether the model successfully recovers the set of target concepts.

\textbf{Edge F1} evaluates local hierarchical correctness by comparing 
predicted parent--child relations against the gold taxonomy. Each directed edge is treated as an evaluation unit. This metric 
captures the accuracy of immediate parent assignments.

\textbf{Ancestor F1} evaluates global hierarchical consistency. For each 
node, we consider its full ancestor set defined by the path from the node 
to the root. Predicted ancestor--descendant pairs are compared against the 
corresponding gold ancestor sets. This metric is particularly sensitive to 
errors in global positioning that Edge F1 alone 
cannot detect.

\paragraph{Baselines.}
We compare \textbf{SPARROW} against three LLM-based taxonomy induction baselines spanning one-pass generation, layer-wise construction, and pairwise scoring.

\begin{itemize}
    \item \textbf{TaxoGPT}~\cite{DBLP:conf/models/ChenYV23} formulates taxonomy induction as a text generation task. Given a set of concepts, the model is prompted in a few-shot setting to generate parent--child relations as textual statements. 
    \item \textbf{Chain-of-Layer (CoL)}~\cite{DBLP:conf/cikm/0001BTFLZ024} constructs taxonomies in a top-down, layer-wise manner. Starting from a root concept, the model iteratively expands the taxonomy by attaching subsets of entities at each level. 
    \item \textbf{LLMscorer} adapts the parent-ranking formulation of LMScorer~\cite{jain-espinosa-anke-2022-distilling}. It uses the backbone model to assess parent--child compatibility within retrieved candidate sets and constructs the taxonomy from the resulting parent assignments. Implementation details are provided in Appendix~\ref{app:implementation}.
\end{itemize}


\subsection{Main Results}

Table~\ref{tab:main_large} presents taxonomy induction results on CCS and Google, each evaluated on a subset of 1{,}000 nodes, under both GPT-5\footnote{\url{https://developers.openai.com/api/docs/models/gpt-5}} and LLaMA3-8B-Instruct\footnote{\url{https://huggingface.co/meta-llama/Meta-Llama-3-8B-Instruct}} backbones.

\paragraph{Results under GPT-5.}
Even with GPT-5, large-scale taxonomy induction remains structurally challenging. SPARROW achieves the strongest global structural quality on both datasets, exceeding TaxoGPT in Ancestor F1 by 0.226 on CCS and 0.126 on Google. LLMscorer attains higher Node and Edge F1, consistent with its independent per-concept parent-selection design, which favors node coverage and locally accurate attachments. SPARROW nevertheless outperforms LLMscorer in Ancestor F1 by 0.147 on CCS and 0.292 on Google. Notably, on Google, this substantial global advantage is achieved with nearly identical Edge F1 (0.579 vs.\ 0.588). These results demonstrate that SPARROW achieves substantially stronger global hierarchical coherence through structure-preserving partitioning and constraint-guided fusion.

\paragraph{Results under LLaMA3-8B-Instruct.}
The limited-context setting is substantially more challenging. TaxoGPT suffers severe recall collapse on both datasets, while Chain-of-Layer cannot complete inference because its layer-wise prompts exceed the model's context window. SPARROW nevertheless achieves the highest Ancestor F1 on both datasets. Relative to TaxoGPT, it achieves approximately $28\times$ higher Node F1 and over $200\times$ higher Ancestor F1 on CCS, as well as over $60\times$ higher Ancestor F1 on Google. Against LLMscorer, SPARROW achieves higher Ancestor F1 on CCS (0.403 vs.\ 0.146) and Google (0.253 vs.\ 0.171), despite lower Node and Edge F1. Notably, SPARROW with LLaMA3-8B-Instruct matches TaxoGPT with GPT-5 on CCS Ancestor F1 (0.403 vs.\ 0.397), despite using a substantially weaker backbone. This cross-backbone comparison further highlights the robustness of SPARROW's divide-and-merge framework under constrained backbone capacity and context.

\begin{figure}[t]
    \centering
    \includegraphics[width=0.95\linewidth]{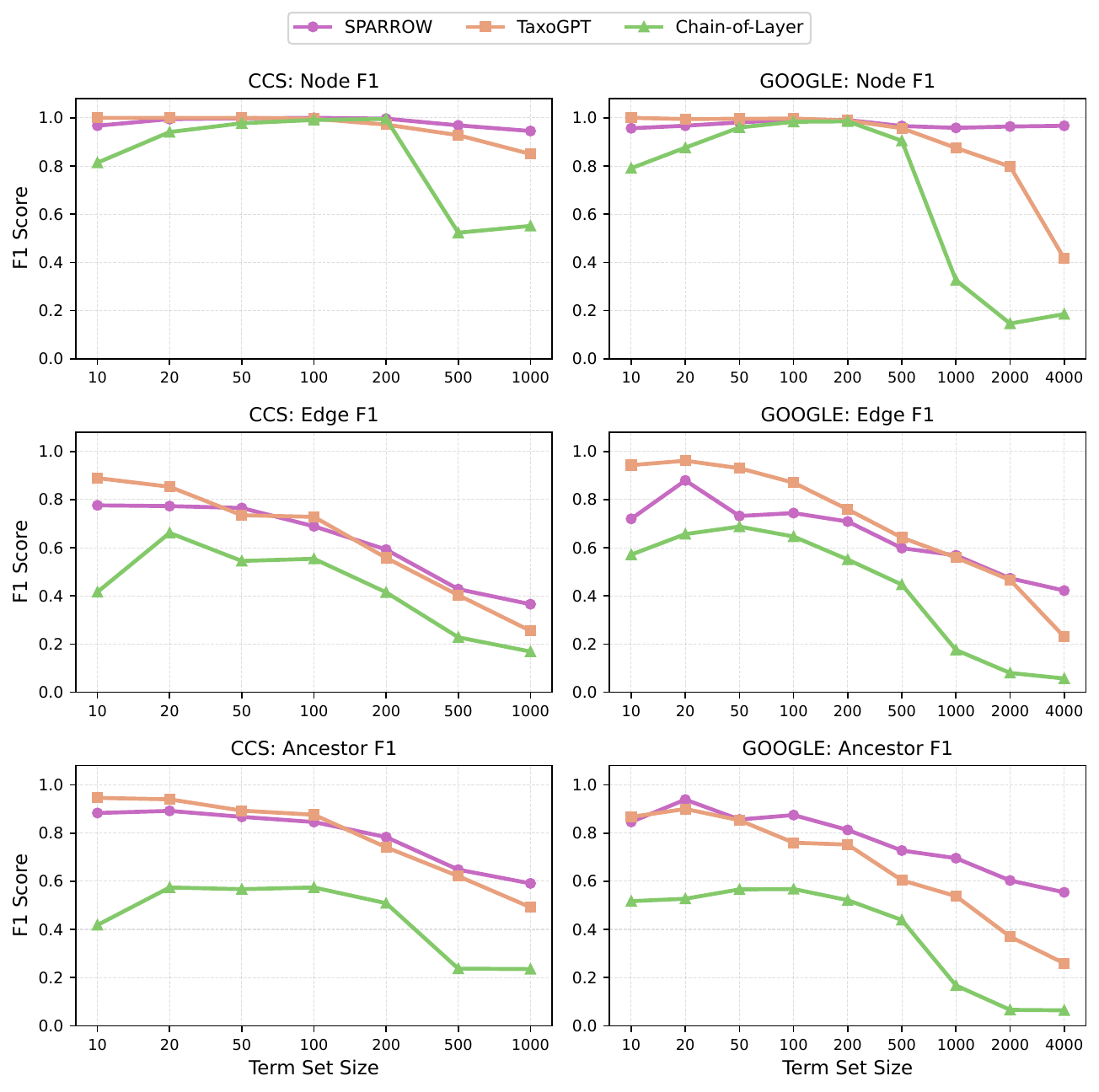}
    \caption{Scalability results on CCS and Google.}
    \label{fig:scalability}
\end{figure}

\subsection{Scalability Analysis}

Figure~\ref{fig:scalability} compares TaxoGPT, Chain-of-Layer, and SPARROW across increasing input scales. As the input size increases, both baselines degrade markedly, whereas SPARROW remains substantially more stable. The resulting performance gap is particularly pronounced in the 4K-node Google setting.

LLMscorer is not included in these curves because, under its per-concept inference protocol, a single sweep over all CCS and Google scales is estimated to consume approximately 4M tokens. Its 1K performance and measured token consumption are reported in Tables~\ref{tab:main_large} and~\ref{tab:cost_analysis}, respectively.

\paragraph{Cross-domain and extreme-scale evaluation.}
We further evaluate SPARROW on SemEval-Food and MeSH (Appendix~\ref{app:additional_results}). SPARROW achieves the highest Node and Ancestor F1 on SemEval-Food and an Ancestor F1 of 0.487 in the 10K-node MeSH setting, compared with 0.011 for TaxoGPT and 0.015 for Chain-of-Layer. These results demonstrate cross-domain generalization and extend our scalability evaluation to 10K concepts.

\subsection{Structure Preservation in the Divide Stage}

\begin{table}[t]
\centering
\small

\setlength{\tabcolsep}{4pt}

\begin{tabular}{lccc}
\toprule
\textbf{Method} & \textbf{Overlap} & \textbf{IER} & \textbf{BID} \\
\midrule

K-means & 0 & 0.620 & 0.133 \\
K-means & $\tau=0.60$ & \textbf{0.902} & 0.116 \\
K-means & $\tau=0.90$ & 0.798 & 0.122 \\

\midrule

Hierarchical & 0 & 0.643 & 0.138 \\

\midrule

Spectral & 0 & 0.766 & \textbf{0.149} \\
Spectral & $\mathrm{thr}=0.25$ & 0.839 & 0.119 \\
Spectral & $\mathrm{thr}=0.05$ & 0.773 & 0.137 \\

\bottomrule
\end{tabular}

\caption{
Comparison of clustering strategies on CCS at the 1K scale.
IER and BID evaluate structure preservation in the divide stage.
}

\label{tab:ccs_clustering_bid}

\end{table}

An effective partitioning strategy should preserve as many ground-truth parent--child relations as possible within blocks while keeping the blocks compact. We quantify 
this with two complementary metrics.\\ \textbf{Intra-block Edge Retention (IER)} 
measures the fraction of ground-truth edges retained within the blocks:
\begin{equation*}
\text{IER} = \frac{\left|\bigcup_{B \in \mathcal{B}} E_{\text{intra}}^{B}\right|}{|E_{\text{gt}}|}.
\end{equation*}
\textbf{Block-normalized Intra-edge Density (BID)} captures structural 
compactness, penalizing partitions that preserve edges only by forming 
excessively large blocks:\\
\begin{equation*}
\text{BID} = \frac{\sum_{B \in \mathcal{B}} |E_{\text{intra}}^{B}|}{\sum_{B \in \mathcal{B}} N_B \log N_B},
\end{equation*}
where $N_B$ is the number of nodes in block $B$.
\begin{table*}[t]
\centering
\small

\setlength{\tabcolsep}{5pt}

\begin{tabular}{lllccccccccc}
\toprule

\textbf{Dataset} & \textbf{Split} & \textbf{Merge}
& \multicolumn{3}{c}{\textbf{Node}}
& \multicolumn{3}{c}{\textbf{Edge}}
& \multicolumn{3}{c}{\textbf{Ancestor}} \\

\cmidrule(lr){4-6}
\cmidrule(lr){7-9}
\cmidrule(lr){10-12}

& & 
& P & R & F1
& P & R & F1
& P & R & F1 \\

\midrule

\multirow{3}{*}{CCS}

& \ding{55} & \ding{55}
& 0.984 & 0.756 & 0.855
& 0.325 & 0.242 & 0.278
& 0.696 & 0.277 & 0.397 \\

& \ding{51} & \ding{55}
& 1.000 & 0.939 & 0.969
& 0.429 & 0.394 & 0.411
& 0.601 & 0.379 & 0.465 \\

& \ding{51} & \ding{51}
& 1.000 & 0.939 & \textbf{0.969}
& 0.425 & 0.399 & \textbf{0.412}
& 0.670 & 0.582 & \textbf{0.623} \\

\midrule

\multirow{3}{*}{Google}

& \ding{55} & \ding{55}
& 1.000 & 0.812 & 0.896
& 0.819 & 0.432 & 0.566
& 0.700 & 0.556 & 0.619 \\

& \ding{51} & \ding{55}
& 1.000 & 0.955 & 0.977
& 0.585 & 0.543 & 0.563
& 0.714 & 0.499 & 0.587 \\

& \ding{51} & \ding{51}
& 1.000 & 0.955 & \textbf{0.977}
& 0.592 & 0.566 & \textbf{0.579}
& 0.770 & 0.721 & \textbf{0.745} \\

\bottomrule

\end{tabular}

\caption{
Ablation study about SPARROW stages on CCS and Google.
}

\label{tab:ablation_datasets}

\end{table*}
\begin{table}[t]
\centering
\small

\setlength{\tabcolsep}{3pt}

\begin{tabular}{ccccc}
\toprule

\textbf{Variant}
& \textbf{Component}
& \textbf{Node F1}
& \textbf{Edge F1}
& \textbf{Anc. F1} \\

\midrule

w/o Scope
& Step 1
& 0.977 & 0.547 & 0.484 \\

Sim.-only
& Step 2
& 0.977 & 0.406 & 0.374 \\

Node-only
& Step 2
& 0.977 & 0.559 & 0.530 \\

Always-sib.
& Step 3
& 0.977 & 0.556 & 0.504 \\

\textbf{SPARROW}
& Full
& 0.977 & \textbf{0.579} & \textbf{0.744} \\

\bottomrule

\end{tabular}

\caption{
Merge-stage ablation study on Google at the 1K scale.
``w/o Scope'' removes constraint-guided candidate scoping (Step 1);
``Sim.-only'' and ``Node-only'' simplify parent selection (Step 2);
``Always-sib.'' disables sibling-or-parent resolution (Step 3).
}

\label{tab:merge_ablation}

\end{table}

\begin{table}[!t]
\centering
\small

\setlength{\tabcolsep}{5pt}

\begin{tabular}{lccc}
\toprule
\textbf{Method} & \textbf{Local} & \textbf{Merge} & \textbf{Total} \\
\midrule

TaxoGPT & 9,813 & -- & 9,813 \\

Chain-of-Layer & 62,038 & -- & 62,038 \\

LLMscorer & 407,518 & -- & 407,518 \\

\textbf{SPARROW} & 10,254 & 52,058 & 62,312 \\

\bottomrule
\end{tabular}

\caption{
API token consumption across different taxonomy induction methods.
For SPARROW, we further decompose the cost into local induction and merge stages.
}

\label{tab:cost_analysis}

\end{table}

As shown in Table~\ref{tab:ccs_clustering_bid}, overlapping K-means with $\tau=0.60$ achieves the highest IER. However, its lower BID suggests that this improvement relies on large, redundant blocks, weakening the scalability benefits of partitioning. In contrast, spectral clustering without overlap achieves the highest BID, producing more compact and structurally efficient partitions. We therefore adopt this method as our partitioning strategy. Appendix~\ref{app:robustness} reports robustness to partitioning noise and sensitivity to block size.

\subsection{Ablation Study}

We conduct an ablation study on CCS and Google at the 1K scale to analyze the contributions of the divide and merge stages in SPARROW as shown in Table~\ref{tab:ablation_datasets}. Removing both stages substantially reduces Edge F1 and Ancestor F1, indicating the difficulty of inducing a coherent hierarchy over a large concept set in a single pass. Introducing the divide stage significantly improves Edge F1 by decomposing the taxonomy into semantically coherent local subgraphs, reducing the input scale and allowing the model to induce parent-child relationships with greater precision within each block. However, Ancestor F1 remains limited, as locally correct subgraphs still lack global positional grounding across blocks. The full SPARROW framework achieves the strongest Ancestor F1 on both datasets, confirming that the divide stage improves local structural fidelity while the merge stage is indispensable for recovering globally coherent hierarchies.

Table~\ref{tab:merge_ablation} further validates the internal design of the merge stage on Google. Without constraint-guided candidate scoping, each node must consider all previously merged nodes as possible parents. This reduces Ancestor F1 by 0.260 relative to SPARROW and uses roughly 6.7~$\times$ more tokens, effectively negating the efficiency gains of the divide stage. For parent selection, Sim.-only and Node-only drop Ancestor F1 by 0.370 and 0.214 respectively compared to SPARROW, confirming that combining embedding-based filtering with path-aware LLM inference is critical for accurate parent resolution. Finally, Always-sib instantiates the \textit{parent displacement} failure mode. By defaulting every attachment to a sibling relation, it restricts attachment points to block root nodes and degenerates into the rigid global fusion shown in Figure~\ref{fig:types}. This reduces Ancestor F1 by 0.240 relative to SPARROW. Together, these results demonstrate that all three components, constraint-guided candidate scoping, path-aware parent selection, and sibling-or-parent resolution, are individually necessary for the merge stage to recover globally coherent hierarchies.

\subsection{Cost Analysis}

Table~\ref{tab:cost_analysis} reports GPT-5 API token consumption on CCS at the 1K scale. TaxoGPT and SPARROW's local stage both rely on one-pass generation and have comparable token usage (9{,}813 vs.\ 10{,}254). Including fusion, SPARROW's total remains nearly identical to Chain-of-Layer (62{,}312 vs.\ 62{,}038 tokens), but its Ancestor F1 is substantially higher (0.623 vs.\ 0.271). Our LLMscorer adaptation replaces the original LMScorer's exhaustive $O(n^2)$ pair scoring with $O(n)$ LLM parent-selection queries over retrieved candidate sets. Even with this reduction, LLMscorer consumes 407{,}518 tokens, approximately $6.5\times$ as many as SPARROW, and achieves lower Ancestor F1 (0.476 vs.\ 0.623). These comparisons show that SPARROW directs computation toward global reconciliation, yielding a stronger accuracy--cost trade-off than methods with comparable or greater token usage.

\section{Conclusion}

We present \textsc{SPARROW}, a scalable divide-and-merge framework 
for large-scale taxonomy induction that explicitly addresses two 
fundamental structural failure modes: structural fragmentation during 
partitioning and parent displacement during merging. Through 
structure-preserving spectral partitioning and constraint-guided 
incremental fusion, SPARROW decomposes large-scale induction into 
tractable local subproblems while preserving global hierarchical 
consistency. Experiments on CCS and Google across both strong- and 
limited-context backbones demonstrate that SPARROW consistently 
achieves superior ancestor-level accuracy and structural stability, 
with gains that become more pronounced at scale. These results suggest that the scalability bottleneck in taxonomy induction is fundamentally structural rather than model-specific, and that principled pipeline design can unlock strong hierarchical reasoning even from limited-capacity backbones.

\section*{Limitations}

While SPARROW achieves strong performance, several limitations remain. SPARROW studies single-parent rooted trees and does not currently support polyhierarchical taxonomies represented as DAGs, in which concepts may have multiple parents. The method also assumes a complete input concept set, whereas important intermediate concepts may be missing in practice, potentially introducing structural bias. Finally, the iterative merge stage incurs token overhead, and improving its efficiency remains future work.

\section*{Acknowledgments}
This research is supported by the Ministry of Education, Singapore, under its MOE AcRF TIER 1 Grant (T1 251RES2517).


\bibliography{custom}
\appendix

\section{Appendix}
\label{sec:appendix}
\subsection{Data Preparation}
\label{app:data_preparation}
\paragraph{Tree Conversion.}
CCS is originally structured as a DAG, where certain concepts appear under multiple parent categories. We convert it to a strict tree via single-parent pruning, retaining only concepts with unambiguous single-parent assignments. This yields 1,768 concepts from the original taxonomy. Google Product Taxonomy is natively tree-structured and requires no conversion.
\paragraph{Sampling Protocol.}
For each target scale, we sample a connected rooted sub-taxonomy by random growth from the original taxonomy root: at each step, we randomly select a node whose parent has already been included in the sampled taxonomy, and then add its children as future expansion candidates. This preserves valid parent--child and ancestor relations inherited from the full taxonomy. To prevent leakage, few-shot examples are sampled from a separate top-level subtree pool, while test subsets are sampled from outside this pool. We evaluate CCS at up to 1K concepts, Google at up to 4K, SemEval-Food at 1K, and MeSH at 1K and 10K. All processed datasets are publicly available at \url{https://github.com/rebeccazyr/SPARROW}.

\subsection{Implementation Details.}
\label{app:implementation}
\paragraph{Model and Prompting Configuration.}
All methods use a 5-shot prompting setting. For TaxoGPT, Chain-of-Layer, and SPARROW, Table~\ref{tab:main_large} reports means over five independent runs, with standard deviations in Table~\ref{tab:standard_deviations}; Figure~\ref{fig:scalability} instead reports one run per scale. LLMscorer outputs are deterministically decycled before evaluation. We use \texttt{GPT-5} via API with default decoding parameters, and deploy LLaMA3-8B locally on NVIDIA H100 GPUs with a temperature of 0.6.

\paragraph{LLMscorer Implementation.}
The original LMScorer scoring paradigm instantiates a natural-language template for each child--parent pair and ranks candidates using likelihood, pseudo-likelihood, or perplexity from a conventional language model. Exhaustively applying this formulation requires scoring all concept pairs, and perplexity is not a natural interface for modern chat models. We therefore implement LLMscorer in two stages. First, concept names are encoded with \texttt{allenai/specter2} \cite{DBLP:conf/emnlp/SinghDCDF23}, and the 50 most similar candidate parents for each child are retrieved by cosine similarity. Second, the child and its candidate list are presented to the evaluated chat backbone, which selects one parent. The resulting edges satisfy the single-parent constraint by construction, and cycles are removed deterministically before evaluation.

\paragraph{Spectral Clustering Configuration.}
For block construction, we apply non-overlapping spectral clustering over entity embeddings produced by \texttt{allenai/specter2}. The number of clusters $K$ is selected automatically via a spectral eigengap heuristic. Specifically, for an entity set of size $n$, we define a backbone-dependent reference count $k^*=\lceil n/s_m\rceil$, where $s_m=70$ for LLaMA3-8B-Instruct and $s_m=100$ for other evaluated backbones. More generally, $s_m$ can be chosen according to the backbone model's scale and reasoning capability. We then search over the candidate range $[\max(2, \lfloor 0.7k^* \rfloor), \min(\lceil 1.3k^* \rceil, n-1)]$, selecting the $K$ that yields the largest eigengap. The affinity matrix is constructed from cosine distances as $\exp(-\gamma d^2)$ with $\gamma=1.0$, using a nearest-neighbor graph with $n\_neighbors=\min(n-1, 10)$.

\paragraph{Taxonomy Fusion Configuration.}
We initialize the global taxonomy using the largest block-level taxonomy and incrementally merge the remaining blocks in descending order of block size. Within each incoming block, nodes are processed in breadth-first order. Parent candidates are retrieved via cosine similarity over already-placed nodes in the \texttt{allenai/specter2} embedding space, retaining the top-$10$ candidates by default. Fusion proceeds in three steps: candidate scoping, parent selection, and sibling-versus-parent resolution.

\setcounter{table}{9}
\begin{table*}[!b]
\centering
\small
\setlength{\tabcolsep}{3.2pt}
\begin{tabular}{@{}rrrcccccccccr@{}}
\toprule
\textbf{Block} & \textbf{\#} & \textbf{IER}
& \multicolumn{3}{c}{\textbf{Node F1}}
& \multicolumn{3}{c}{\textbf{Edge F1}}
& \multicolumn{3}{c}{\textbf{Ancestor F1}}
& \textbf{Fusion} \\
\textbf{size} & \textbf{blocks} &
& \textbf{Local} & \textbf{Fused} & $\boldsymbol{\Delta}$
& \textbf{Local} & \textbf{Fused} & $\boldsymbol{\Delta}$
& \textbf{Local} & \textbf{Fused} & $\boldsymbol{\Delta}$
& \textbf{tokens} \\
\cmidrule(lr){4-6}\cmidrule(lr){7-9}\cmidrule(lr){10-12}
1000 & 1  & 1.000 & 0.855 & 0.855 & 0.000 & 0.278 & 0.278 & 0.000 & 0.397 & 0.397 & 0.000 & 0 \\
500  & 2  & 0.845 & 0.940 & 0.940 & 0.000 & 0.365 & 0.367 & +0.002 & 0.620 & 0.640 & +0.020 & 25,862 \\
100  & 10 & 0.766 & 0.969 & 0.969 & 0.000 & 0.411 & 0.412 & +0.001 & 0.465 & 0.623 & +0.158 & 52,058 \\
50   & 20 & 0.587 & 0.997 & 0.997 & 0.000 & 0.369 & 0.379 & +0.010 & 0.319 & 0.524 & +0.205 & 60,471 \\
\bottomrule
\end{tabular}
\caption{Block-size sensitivity on CCS at 1K scale. $\Delta$ is the F1 change from the local block taxonomies to the fused taxonomy; size 100 balances local induction, edge retention, and fusion cost.}
\label{tab:block_sensitivity}
\end{table*}

\setcounter{table}{5}

\subsection{Additional Experimental Results}
\label{app:additional_results}

\begin{table}[t]
\centering
\small
\setlength{\tabcolsep}{3pt}
\resizebox{\columnwidth}{!}{%
\begin{tabular}{lrlccc}
\toprule
\textbf{Dataset} & \textbf{$n$} & \textbf{Method} & \textbf{Node} & \textbf{Edge} & \textbf{Anc.} \\
\midrule
\multirow{3}{*}{SemEval-Food} & \multirow{3}{*}{1K}
& TaxoGPT & 0.810 & \textbf{0.606} & 0.567 \\
& & Chain-of-Layer & 0.513 & 0.315 & 0.227 \\
& & \textbf{SPARROW} & \textbf{0.965} & 0.475 & \textbf{0.595} \\
\midrule
\multirow{6}{*}{MeSH} & \multirow{3}{*}{1K}
& TaxoGPT & 0.857 & \textbf{0.547} & 0.658 \\
& & Chain-of-Layer & 0.271 & 0.113 & 0.166 \\
& & \textbf{SPARROW} & \textbf{0.964} & 0.540 & \textbf{0.700} \\
\cmidrule(lr){2-6}
& \multirow{3}{*}{10K}
& TaxoGPT & 0.102 & 0.030 & 0.011 \\
& & Chain-of-Layer & 0.040 & 0.010 & 0.015 \\
& & \textbf{SPARROW} & \textbf{0.902} & \textbf{0.360} & \textbf{0.487} \\
\bottomrule
\end{tabular}}
\caption{Cross-domain and extreme-scale F1 results with GPT-5. Bold marks the best value for each dataset and scale.}
\label{tab:additional_datasets}
\end{table}

\begin{table}[t]
\centering
\small
\setlength{\tabcolsep}{3pt}
\resizebox{\columnwidth}{!}{%
\begin{tabular}{llccc}
\toprule
\textbf{Backbone} & \textbf{Method} & \textbf{Node} & \textbf{Edge} & \textbf{Anc.} \\
\midrule
\multirow{3}{*}{LLaMA3.3-70B}
& TaxoGPT & 0.042 & 0.015 & 0.018 \\
& Chain-of-Layer & 0.093 & 0.034 & 0.052 \\
& \textbf{SPARROW} & \textbf{0.499} & \textbf{0.132} & \textbf{0.231} \\
\midrule
\multirow{3}{*}{gpt-oss-120b}
& TaxoGPT & 0.268 & 0.064 & 0.115 \\
& Chain-of-Layer & 0.292 & 0.003 & 0.092 \\
& \textbf{SPARROW} & \textbf{0.682} & \textbf{0.161} & \textbf{0.312} \\
\bottomrule
\end{tabular}}
\caption{F1 results on CCS at the 1K scale with additional open-weight backbones.}
\label{tab:additional_backbones}
\end{table}

\begin{table}[t]
\centering
\small
\setlength{\tabcolsep}{4pt}
\begin{tabular}{llccc}
\toprule
\textbf{Dataset} & \textbf{Method} & \textbf{Node} & \textbf{Edge} & \textbf{Anc.} \\
\midrule
\multirow{3}{*}{CCS}
& TaxoGPT & \textbf{0.024} & 0.104 & 0.062 \\
& Chain-of-Layer & 0.285 & 0.077 & 0.117 \\
& \textbf{SPARROW} & \textbf{0.024} & \textbf{0.032} & \textbf{0.035} \\
\midrule
\multirow{3}{*}{Google}
& TaxoGPT & 0.031 & 0.120 & 0.082 \\
& Chain-of-Layer & 0.339 & 0.194 & 0.185 \\
& \textbf{SPARROW} & \textbf{0.024} & \textbf{0.045} & \textbf{0.045} \\
\bottomrule
\end{tabular}
\caption{Standard deviations of GPT-5 F1 results over five runs. Bold marks the lowest deviation.}
\label{tab:standard_deviations}
\end{table}

\begin{table}[t]
\centering
\small
\setlength{\tabcolsep}{3pt}
\resizebox{\columnwidth}{!}{%
\begin{tabular}{rcccc}
\toprule
\textbf{CER (\%)} & $\Delta$\textbf{Node F1} & $\Delta$\textbf{Edge F1} & $\Delta$\textbf{Anc. R} & $\Delta$\textbf{Anc. F1} \\
\midrule
11.6 & 0.000 & $-$0.011 & +0.190 & +0.148 \\
30.0 & 0.000 & +0.007 & +0.135 & +0.095 \\
50.0 & 0.000 & +0.017 & +0.135 & +0.088 \\
\midrule
Avg. & 0.000 & +0.004 & +0.153 & +0.110 \\
\bottomrule
\end{tabular}}
\caption{Gain from constraint-guided fusion over unfused block taxonomies under controlled partitioning noise. CER denotes Cut Edge Ratio.}
\label{tab:noise_robustness}
\end{table}

Table~\ref{tab:additional_datasets} extends the GPT-5 evaluation beyond the two primary benchmarks. On SemEval-Food, SPARROW achieves the best Node and Ancestor F1, while TaxoGPT obtains the best Edge F1. On MeSH, SPARROW achieves the best Ancestor F1 at both 1K and 10K concepts and retains substantially stronger global structure at 10K. Table~\ref{tab:additional_backbones} shows the same relative advantage with two additional open-weight backbones on CCS. Table~\ref{tab:standard_deviations} reports variability across the five GPT-5 runs used in the main table.

\subsection{Robustness and Block-Size Sensitivity}
\label{app:robustness}

To isolate robustness to partitioning errors, we introduce controlled cut edges and compare the fused result with the corresponding unfused block taxonomies. Table~\ref{tab:noise_robustness} shows that fusion preserves a positive Ancestor F1 gain even at a 50\% Cut Edge Ratio (CER), indicating that the merge stage can recover global paths despite substantial fragmentation. Table~\ref{tab:block_sensitivity} varies the reference block size on CCS at 1K. Smaller blocks increase the number of cut edges and fusion tokens but also enlarge the recoverable Ancestor F1 gain; size 100 provides the best balance in our setting.

\subsection{Embedding-Space Connectivity}
\label{app:embedding_connectivity}

\setcounter{table}{10}
\begin{table}[t]
\centering
\small
\setlength{\tabcolsep}{4pt}
\begin{tabular}{lcc}
\toprule
\textbf{Dataset} & \textbf{Connectivity} & \textbf{Centroid purity} \\
\midrule
CCS & 90.8 / 13.9 & 41.0 / 3.3 \\
Google & 92.2 / 14.0 & 50.8 / 3.9 \\
\bottomrule
\end{tabular}
\caption{Embedding-space structure of gold subtrees versus size-matched random sets (gold/random, \%). Connectivity is measured in the SPECTER2 $k$NN graph; centroid purity measures concentration around a shared center.}
\label{tab:embedding_connectivity}
\end{table}

We compare gold subtrees with size-matched random concept sets in the \texttt{allenai/specter2} embedding space. Connectivity is the fraction of nodes belonging to the largest connected component of the induced $k$NN graph, and centroid purity measures concentration around the set centroid. Table~\ref{tab:embedding_connectivity} shows that gold subtrees are highly connected but only moderately centroid-compact. This supports graph-based partitioning while avoiding the stronger assumption that each subtree forms a compact convex cluster.


\subsection{Qualitative Fusion Cases}

\paragraph{Recovering fragmentation.}
The local edge between \emph{Data structures design and analysis} and its parent \emph{Design and analysis of algorithms} is cut across blocks. Constraint-guided fusion retrieves the latter as the parent and restores the missing cross-block relation.

\paragraph{Correcting parent displacement.}
For the path \emph{Computer vision} $\rightarrow$ \emph{Computer vision tasks} $\rightarrow$ \emph{Scene understanding}, treating the query as a sibling would flatten the hierarchy. Sibling-or-parent resolution instead places it below the selected node, preserving the intermediate level.

\paragraph{Unrecovered spurious parent.}
For \emph{Spam detection}, whose gold parent is \emph{Web search engines}, the system selects \emph{Software and application security}. This semantically plausible but structurally incorrect attachment scopes the concept to the wrong subtree and illustrates the remaining sensitivity to candidate retrieval and local semantic ambiguity.

\raggedbottom
\subsection{Prompts for SPARROW}

\noindent
\begin{tcolorbox}[colback=gray!5, colframe=gray!50!black, title=Block-Level Taxonomy Induction]
\small
\label{direct_llm_taxonomy}

\textbf{Input:}

\vspace{4pt}
\textbf{Concept set $\mathcal{C}$}: a flat list of concepts.

\vspace{4pt}
\textbf{Few-shot examples}: concept sets paired with reference parent--child relations.

\vspace{4pt}
\texttt{==========}

\vspace{4pt}
\textbf{Instruction:}

\vspace{4pt}
Generate parent--child relations that organize the input concepts into a taxonomy hierarchy.

\vspace{4pt}
\texttt{==========}

\vspace{4pt}
\textbf{Output:}

\vspace{4pt}
A list of parent--child relations in the form:

\vspace{4pt}
\texttt{child is a subtopic of parent}

\end{tcolorbox}

\phantomsection
\label{box:parent_selection}
\noindent \begin{tcolorbox}[colback=gray!5, colframe=gray!50!black, title=Parent Selection] \small 
\label{parent_selection}
\textbf{Input:} 

\vspace{4pt} \textbf{Query concept $Q$} (concept label, child concepts, and local taxonomy path hint)

\vspace{4pt} \textbf{Candidate parent set $\mathcal{P}$} (candidate labels, current child concepts and existing taxonomy paths)

\vspace{4pt} \texttt{==========} \vspace{4pt}

\textbf{Decision criteria:}

\vspace{4pt} \textbf{1. Subsumption}: $P$ is valid only if every child of $Q$ can be semantically placed under $P$.

\vspace{4pt} \textbf{2. Granularity alignment}: Among candidates passing subsumption, prefer $P$ whose existing children are at a similar or slightly broader abstraction level than $Q$'s children.

\vspace{4pt} \textbf{3. Depth continuity}: Prefer attachments that preserve smooth hierarchical progression and avoid skipping semantic levels.

\vspace{4pt} \textbf{Fallback:} If no non-root candidate meet requirements, select ROOT.

\vspace{4pt} \texttt{==========} \vspace{4pt}

\textbf{Output:} 

\vspace{4pt} The label of the selected parent, or ROOT \end{tcolorbox}

\vspace{6pt}
\phantomsection
\noindent \begin{tcolorbox}[colback=gray!5, colframe=gray!50!black, title=Sibling-or-Parent Resolution] \small 
\label{Sibling-or-Parent Resolution}
\textbf{Input:} 

\vspace{4pt} \textbf{Query concept $Q$} (concept label, child concepts, and local taxonomy path hint)

\vspace{4pt} \textbf{selected parent $P$} (candidate labels, current child concepts and existing taxonomy paths)

\vspace{4pt} \texttt{==========} \vspace{4pt}

\textbf{Decision options:}

\vspace{4pt} \textbf{Sibling}: $Q$ is attached as a direct child of $P$, if $Q$'s scope is comparable in granularity to the existing children of $P$.

\vspace{4pt} \textbf{Parent}: $Q$ is inserted as an intermediate node between $P$ and a subset of its existing children, if $Q$ semantically subsumes that subset. The affected children are explicitly identified and moved under $Q$.

\vspace{4pt} \texttt{==========} \vspace{4pt}

\textbf{Output:} 

\vspace{4pt} The relationship type (sibling or parent), and if parent, the list of affected children to be moved under $Q$. \end{tcolorbox}
\end{document}